\documentclass[runningheads]{llncs}
\usepackage{balance}
\usepackage{floatrow}
\usepackage[rightcaption]{sidecap}

\usepackage[show]{notes-alt}
\usepackage{latexsym}

\usepackage{balance}

\PassOptionsToPackage{table}{xcolor} %
\usepackage{hyphenat}
\usepackage{url}
\usepackage{xspace} %
\usepackage{amsmath}
\usepackage{mathtools}
\usepackage{wraptable}
\usepackage{amsfonts}
\usepackage{algorithm}
\usepackage[noend]{algpseudocode}

\usepackage{chngcntr} %
\usepackage[table]{xcolor}
\usepackage{comment}
\usepackage{balance}
\usepackage{enumitem}
\usepackage{soul}
\usepackage{tikz}
\usepackage{tabularx}
\usetikzlibrary{fit,positioning,calc,shapes}
\usepackage{pgfplots}
\pgfplotsset{compat=1.13}
\usepgfplotslibrary{colorbrewer}
\usepackage{multirow}
\usepackage{breqn}
\usepackage{caption}
\usepackage{pifont}
\usepackage{stfloats}
\usepackage{booktabs}
\usepackage{subcaption}
\usepackage{stmaryrd}

\usepackage{floatrow}
\newfloatcommand{capbtabbox}{table}[][\FBwidth]

\DeclareMathOperator*{\argmax}{\arg{}\,\max}

\definecolor{cycle1}{RGB}{228, 26, 28}
\definecolor{cycle2}{RGB}{55, 126, 184}
\definecolor{cycle3}{RGB}{77, 175, 74}
\definecolor{cycle4}{RGB}{152, 78, 163}
\definecolor{cycle5}{RGB}{255, 127, 0}
\definecolor{cycle6}{RGB}{153, 153, 153}%
\definecolor{cycle7}{RGB}{166, 86, 40}
\definecolor{cycle8}{RGB}{247, 129, 191}

\newcommand{\cmark}{\textcolor{cycle3}{\ding{52}}} %
\newcommand{\xmark}{\textcolor{cycle1}{\ding{56}}}
\newcommand{\mehmark}{\mbox{\cmark\textsubscript{\kern-0.45em\tiny\xmark}}}

\newcommand{\mpara}[1]{\medskip\noindent{\bf #1}}

\renewcommand{\P}{\mathcal{P}}

\newcommand{\fromto}{\longrightarrow}
\renewcommand{\to}{\fromto}

\newcommand{\greedy}{\textsc{greedy} }
\newcommand{\greedycov}{\textsc{greedy-cover} }
\newcommand{\greedycovep}{\textsc{greedy-cover-$\epsilon$} }

\DeclarePairedDelimiterX{\norm}[1]{\lVert}{\rVert}{#1}

\newcommand{\mslr}{\texttt{MSLR}}

\newcommand{\mq}{\texttt{MQ2008}}

\newcommand{\pow}{\texttt{PointWise}}
\newcommand{\paw}{\texttt{PairWise}}
\newcommand{\liw}{\texttt{ListWise}}

\newcommand{\sone}{\textsc{Shap-1}}
\newcommand{\lime}{\textsc{Lime}}
\newcommand{\sfive}{\textsc{Shap-5}}
\newcommand{\shap}{\textsc{Shap}}
\newcommand{\ltr}{\textsc{LtR}}

\newcommand{\q}{\mathbf{q}} 
\newcommand{\x}{\boldsymbol{x}}

\newcommand{\f}{\mathcal{F}}

\providecommand{\norm}[1]{\lVert#1\rVert}

\date{}

\begin{document}

\title{Valid Explanations for Learning to Rank Models}

\author{Jaspreet Singh\inst{1} \and
Zhenye Wang\inst{2} \and
Megha Khosla\inst{1} \and
Avishek Anand\inst{1,2}}

\authorrunning{Singh, Wang, Khosla, and Anand}

\institute{L3S Research Center, Hannover, Germany. \and
Leibniz Universit\"at Hannover, Appelstrasse 4, 30167 Hannover, Germany.\\
\email{\{singh,khosla,anand\}@l3s.de, bismarckderzweit@gmail.com}}

\maketitle

\begin{abstract}
Learning-to-rank (LTR) is a class of supervised learning techniques that apply to ranking problems dealing with a large number of features.

The popularity and widespread application of LTR models in prioritizing information in a variety of domains makes their scrutability vital in today's landscape of fair and transparent learning systems. 
However, limited work exists that deals with interpreting the decisions of learning systems that output rankings. In this paper we propose a model agnostic local explanation method that seeks to identify a small subset of input features as explanation to a ranking decision.
We introduce new notions of validity and completeness of explanations specifically for rankings, based on the presence or absence of selected features, as a way of measuring goodness. 
We devise a novel optimization problem to maximize validity directly and propose greedy algorithms as solutions.
In extensive quantitative experiments we show that our  approach outperforms other model agnostic explanation approaches across pointwise, pairwise and listwise LTR models in validity while not compromising on completeness. 

\end{abstract}





\section{Introduction}
\label{sec:intro}


Interpreting the reasons behind predictions made by modern machine learning models is an important yet challenging problem. As a result, many approaches have been proposed to tackle the problem of interpretability for several model families -- decision trees, sequential models, convolutional networks -- and problem settings -- text classification, image recognition, and regression tasks.
However, limited work has been done for interpreting the ranking task and specifically learning to rank (\ltr{}) models that are widely utilized in search and recommendation systems.
Initial work on ranking models has been limited to the ad-hoc retrieval scenario where the complex ranking models under consideration focused purely on better text representation~\cite{singh2020model,Fernando:2019:SIN:3331184.3331312,Singhexs2019,Verma2019}.
However, in many practical scenarios like web search, recommender systems, vertical search, etc. ranking is typically performed using \ltr{} models operating on a large set of features of which text-based features are but a small subset. 

Typical \ltr{} models are trained using standard ML algorithms adjusted to a ranking objective. 
While explanation approaches exist for standard ML models like neural networks~\cite{montavon2017explaining} and decision trees~\cite{alvarez2004explaining}, they are \textit{model introspective}. Such explanation methods are inextricably tailored towards specific model families. 
Consequently \textit{model agnostic} approaches, like \shap~\cite{lundberg2017unified} and \lime{}~\cite{ribeiro2016should} have been popularised that treat the trained model as a functional black box.
However, most of these  approaches have been devised specifically for classification and regression problems and to the best of our knowledge, no approach exists for explaining ranked outputs of \ltr{} models. 

Rankings make for an inherently challenging scenario unlike classical classification and regression tasks in that rankings can be viewed as aggregations of multiple predictions -- sorting regression scores (\pow{} \ltr{}), aggregating preferences (\paw{} \ltr) or choosing permutations (\liw{} \ltr{}). Furthermore, explanations for rankings should not only explain single item relevance but more importantly be able to accurately describe rank differences between items post aggregation.

In this paper we propose interpretability methods for \ltr{} models that enable us to explain entire ranked outputs rather than explaining a single item score or a  classification result. Secondly, we propose a model agnostic feature attribution/explanation method that can help us cross-examine various ranking algorithms coherently without having to worry about the training regimen (pointwise, pairwise, listwise~\cite{liu2009learning}), access to training data and access to the model parameters. 
Such approaches are not only helpful when debugging for model developers but especially so for audits and legislative procedure where a balance needs to be struck between exposing business secrets and transparency. 


\mpara{Our Contributions} 
We propose the problem setting of explaining \ltr{} models as: given a trained ranking model and a ranking produced for a query, our aim is to select a \textit{subset of features} as an explanation that is most responsible for the prediction.

Our first contribution is to define feature explanations of a given pre-trained \ltr{} model in terms of their \textit{validity} and \textit{completeness}. \textit{Validity} encodes the amount of predictive capacity contained in the explanatory features measured in terms of rank correlation with the original ranking. 
\textit{Completeness} on the other hand measures the lack of information in the features that are not explanation features.
Note that a feature subset might be valid but not complete. 
Consider an example where the two variables encode the predictive capacity for a ranking decision and are perfectly correlated. In such a scenario, selecting any one of these variables would result in high validity. However, if only of these features is chosen in the explanation set, the non-explanation set also retains enough predictive capacity. Validity and completeness measures have been proposed earlier ~\cite{deyoung2019eraser} but we are the first in our knowledge to propose an algorithms to explicitly optimize it for \ltr{}.

As our second contribution, we pose the problem of explaining rankings in terms of a size-bounded feature subset that has maximum validity or rank correlation with the original ranking. We propose and empirically analyse greedy heuristics to find such valid explanation sets. 







As a final contribution, we carry out extensive experimental evaluation where we show that our proposed approach is significantly better than \shap{} (between 10\% and 100\%) for finding explanations that are valid yet complete for \ltr{} models.
Specifically, we devise 3 carefully considered greedy heuristics and empirically compare them against the popular \shap{} on a variety of \ltr{} models and 2 commonly used \ltr{} datasets -- \mq{} and \mslr{}~\cite{QinL13}.

\section{Related Work }
\label{sec:related-work}

A large family of methods for posthoc interpretability of ML models have been proposed in recent literature~\cite{guidotti2018survey}.
Broadly speaking, we classify the methods of posthoc interpretability into model introspective methods, where we have access to the model parameters, and model agnostic methods where we use the trained model as a functional blackbox.
Model introspective methods typically attribute importance to input features by model specific optimizations like gradient flows~\cite{montavon2017explaining}, attention values~\cite{captioningxu2015showattention} etc.

Model agnostic methods, popularized by~\cite{ribeiro2016should,alvarez-melis-jaakkola-2017-causal,guidotti2019factual}, on the other hand are more general in that they are agnostic to how the underlying model was trained or how the scores are computed.
Most model agnostic methods formulate simple proxy models attempting to learn a locally faithful approximation around the prediction, for example through linear models or sets of rules, representing sufficient conditions on the prediction~\cite{rulesletham2015interpretable} or by using linear models over local perturbations~\cite{ribeiro2016should,guidotti2019factual}. 
In this work, we operate in the \textbf{model agnostic} regime where we do not assume any access to the ranking model's parameters.  

Beyond these approaches, there are game-theoretic approaches like \textsc{Shap}~\cite{lundberg2017unified} that use model agnostic or introspective approaches as kernels while preserving important properties for explanation like accuracy, missingness and consistency.
 \textit{Shapley Additive Explanations} computes feature importance or contribution values to a prediction (for a specific input) as compared to its expected value. 
In Section~\ref{sec:compareSHAP} we describe the adaptation and limitations of \shap{} for explaining ranking models.
In our experiments, we extend Kernel \shap{} -- a model agnostic variant to estimate shapley values~\cite{lundberg2017unified}, for ranking models and use it as a baseline. Note that Kernel \shap{} is similar to the popular \lime{} approach~\cite{ribeiro2016should} as stated in~\cite{lundberg2017unified}.

Ranking models, like \ltr{}, are different from the classical prediction tasks in that their output is an ordering of input items. This ordering can be obtained directly as model output or as an aggregation over multiple individual ranking decisions.
Consequently, none of the existing work on posthoc interpretability can be directly applied to our setting.
Model agnostic approaches for interpretability in rankings include \cite{singh2018posthoc} which constructed surrogate models to imitate decisions of black-box \ltr{} models. 
Other works~\cite{Singhexs2019,Verma2019} adapt an existing local linear explanation model, \lime{}\cite{ribeiro2016should} to the problem of explaining textual document relevance with respect to a query. 
Both approaches deliver term based explanations for pointwise models and their applicability beyond text features is limited.
We, on the other hand, focus on explaining \ltr{} models with arbitrary features and can be applied additionally to pairwise and listwise approaches. 
Another line of work \cite{singh2020model} proposes an approach for locally explaining the ranked list output of pure text based neural rankers (that learn latent feature representations) via a set of intent terms. Our problem setting however is the more commonly prevailing setting in \ltr{} where we have a known set of input features such as TF-IDF scores, page rank, etc. from which to choose an explanation.

\section{Explaining Rankings}
\label{sec:problem}

In this work, we focus on the ranking task in a document retrieval setting where given a query $\q$ the trained \ltr{} model outputs a ranking of documents that is most relevant to $\q$. We are then interested in explaining the decision of the trained \ltr{} model for a given output ranking with respect to the query $\q$.

\ltr{} models are trained either in a \pow{}, \paw{}, or \liw{} manner~\cite{liu2009learning}. In a \pow{} approach, the \ltr{} problem is cast into a regression problem where the goal is to predict a ranking score for a query-document pair. 
 The \paw{} approach, on the other hand, trains a binary classifier model that classifies the document preference pair as concordant or discordant. 
 In particular, the model outputs the likelihood or probability of the the input pair being a concordant pair. We utilize these output probabilities for the input ordered document pairs as scores in our approach.
Finally, \liw{} approaches  directly optimize the value of ranking evaluation measures, like MAP, NDCG, averaged over all queries in the training data. 
Typical features used in \ltr{} include the frequencies of the query terms in the document, the BM25 and PageRank scores. In our setting we are independent of the training regimen since we treat the \ltr{} model as a blackbox.

\subsection{Explanation Definition}

Given a trained \ltr{} model $\mathcal{R}$ and an input feature space $\f$, we are interested in explaining the output ranking $\mathcal{R}( \q, \f) \to \pi$,  for $\q$ in terms of a subset of features $\f' \subseteq \f$ that are most \textit{impacting} for $\pi$. $\q$ here is shorthand for all retrieved documents for a given query.

We define two measures, namely \emph{validity} and \emph{completeness}, to quantify the importance of a feature subset (explanation) for $\pi$. We use Kendall's tau rank correlation coefficient to quantify the validity and the completeness of our explanations which is defined over a pair of rankings $\pi,\pi'$ as 
$$ \tau(\pi,\pi') = {\text{no. of concordant pairs}- \text{no. of discordant pairs} \over \text{total no. of pairs}}. $$
\begin{definition}[Validity] An explanation $\f' \subset \f$ for $\pi$ is said to be valid if $\mathcal{R}(\q, \f') \to \pi' \approx \pi$, i.e., the model's output is predictable from the explanation alone. In our setting a valid explanation implies that the returned smaller set of features are sufficient to reconstruct the original ranking output by the model (when it used all the features). In particular we measure \textit{validity score} ($\mathcal{V}$) of our explanation by computing the rank correlation between the ranking returned by the model when using the explanation and the original ranking.
$$\mathcal{V}= \tau(\mathcal{R}(\q, \f'), \mathcal{R}(\q, \f)).$$
\end{definition}
\begin{definition}[Completeness]An explanation $\f' \subset \f $ is said to be complete with respect to a trained model $\mathcal{R}$ and $\q$ if removing or altering the explanation features from the input will change the output function or ranking considerably. In other words, an explanation $\f'$ is complete if  $\mathcal{R}(\q, \f \setminus \f')$ cannot approximate $\pi$. We define the completeness score of an explanation as the negative of the rank correlation between the original ranking and the ranking obtained by using features which are not included in the explanation, i.e., 
$$ \mathcal{C}= -\tau(\mathcal{R}(\q, \f \setminus \f'), \mathcal{R}(\q, \f)).$$

\end{definition}
Intuitively, a feature subset $\f'$ is \textit{valid} if it contains enough information to re-create $\pi$ when $\f \setminus \f'$ is not considered for ranking.
Alternately $\f'$ is \textit{complete} if $\f \setminus \f'$ does not contain enough information capacity to re-create the ranking $\pi$. Note that a valid feature subset $\f'$ might not always be complete. This case arises, for example, when two or more predictive features are correlated -- only one of them suffices to be in the valid set but both are required for the complete set. In the following section we formulate the optimization problem for finding explanations and describe our approach.
\subsection{Finding Fixed Size Valid Explanations}
\label{sec:approach}
Validity and completeness give us two ways of measuring the impact of an explanation for rankings. By defining them using $\tau$, we have direct measures that we can optimize for. To find the ideal valid or complete feature attribution, one must exhaustively search all possible feature subsets $\f'$ of all possible sizes that maximize the respective scores. Trivially, when $\f' = \f$, then $\f'$ has maximum validity and completeness, i.e., the entire feature set $\f$ is a valid and complete explanation. 

We emphasize that any human understandable explanation should be small in size. Therefore we restrict the size of explanations to a small constant $k<< |\f|$. Mathematically, if we are interested in finding a \emph{valid} explanation $\f'$ such that $|\f'|=k$, we solve the following optimization problem:

\begin{equation}
\label{eq:validity}
    \argmax_{\f'} \tau(\mathcal{R}(\q, \f'), \mathcal{R}(\q, \f))\,\,\,\,\,\,\,\,\, s.t.\,\,\, |\f'| = k
\end{equation}



Note that one can optimize for either validity or completeness to find an explanation but we choose to focus on validity in this paper. While the measures and optimization formulations may seem relatively similar there is a subtle point of contention. Valid explanations explain the construction of a ranked list whereas completeness in essence is a measure of sensitivity. Considering that we want a fixed size explanation, valid explanations inherently mean small succinct sets whereas complete explanations imply larger feature sets since they must take all feature correlations into account.



\mpara{Theoretical Properties and Limitations.} Before we describe our approach, let us examine some theoretical constraints of the optimization objective. The naive approach to solve equation~\ref{eq:validity} is to compute $\tau$ for all possible k-sized subsets. This is an exact solution but is computationally infeasible when $n >> k$ where $n=|\f|$. In practice $n \geq 100$ and $k \leq 10$ usually which means that we have ${n \choose k} = {100 \choose 10} \approx 1.7e^{13}$ possibilities. Hence we must find a way to prune the $\f'$ search space while still trying to find the optimal solution. A commonly used technique in such a setting is greedy incremental search as shown in Algorithm~\ref{alg:incSearch}. In case the optimization function  is  submodular or even weakly submodular \cite{submodularity2011}, greedy selection gives us a theoretical bound on the error of our selection. The (weak) submodularity condition can be checked using the following definition from \cite{submodularity2011}.
\begin{definition}(Submodularity Ratio) Let $g$ be a nonnegative set function. The submodularity ratio of $g$ with respect to a set $U$ and a parameter $k \geq 1$ is $$\gamma_{U, k}(g)=\min _{L \subseteq U, S:|S| \leq k, S \cap L=\emptyset} \frac{\sum_{x \in S} g(L \cup\{x\})-g(L)}{g(L \cup S)-g(L)},$$
where the ratio is considered to be equal to $1$
when its numerator and denominator are both
$0$. Then $g$ is submodular if and only if $\gamma_{U, k}(f)\ge 1$ for all $U$ and $k$. If $\gamma = \min_{U,k} \gamma_{U, k}(f) \in (0,1)$, the function is said to be weakly submodular.
\end{definition}
In our case $U$ is the set of all input features, $L$ and $S$ are non-intersecting subsets of $U$ and the function $g$ (for a particular ranking model) corresponds to the rank correlation between the original ranking and a new ranking obtained by adding new feature to an existing explanation set. Because of the black box nature of our ranking models, neither submodularity nor weak submodularity of the objective function can be guaranteed which hinders us from achieving any theoretical guarantees for the solution. Nevertheless, in practice, greedy solutions have been shown to provide convincing and better results (for example see \cite{streamweaksub2017}). In this work, we propose variants of a greedy approach to find valid explanations and empirically analyze the effectiveness of our proposed methods.

\subsection{Our Solution Framework}
\label{sec:matrix}


In greedy incremental search, we begin with an empty explanation set and add features incrementally as long as the objective criteria is non-decreasing. Rather than optimizing directly for $\tau$ we optimize for a proxy utility measure based on the relevance score difference instead of rank difference between documents. The justification for which will become clearer in the remainder of this section. In the following we elaborate on our proxy measure and our greedy selection procedure.


\mpara{Step 1: Preference Matrix.} We first construct a preference matrix where the rows represent features or subsets of features from $\f$ and the columns are concordant pairs of documents as determined by the original ranking $\pi$. For short ranked lists we use all pairs but for longer lists we sample uniformly at random from all concordant pairs as determined by $\pi$. Please note that as the top ranked documents appear in a larger number of concordant pairs, they will be more likely to appear in the sampled columns. Let $P$ denote the set of sampled concordant pairs from $\pi$. 



\mpara{Step 2: Propensity calculation.} Each row in our matrix is meant to denote the impact of adding $f$ to $\f'$ for every $p \in P$. By observing which pairs are concordant when adding $f$ we can directly optimize for $\tau$. A simple binary indicator of concordance may at first glance seem to be ideal but in the greedy setting purely looking at ranks can be volatile. Adding a new feature to $\f'$ that does not increase pairwise concordance but changes the relevance scores is also important to consider. \textbf{Firstly} (and trivially), a positive score difference directly implies concordance. \textbf{Secondly}, by observing the magnitude of score difference we can start to estimate the likelihood of the pair staying concordant (or becoming concordant) as more features are added to $\f'$ thereby introducing a level of robustness lacking when only looking at rank differences incrementally when directly optimizing for $\tau$. 


Hence we compute each cell value $z^{f}_{p}$ of the preference matrix as the likelihood of the pair of being concordant when feature $f$ is added to the explanation set instead of a simple binary indicator. If $z^{f}_{p} > 0$, $p$ is concordant. Formally, if $\mathcal{R}( \x_{i},\f')$ denotes the score for document $\x_{i}$ when only features in $\f'$ were used to represent the input to an already trained \ltr{} model while masking all features in $\f\setminus \f'$. Then we compute 
\begin{equation}
\label{eq:rowimpact}
     z^{f}_{p} = (\mathcal{R}( \x_{i},\f' \cup f) - \mathcal{R}(\x_{j},\f' \cup f)) \cdot w_p
\end{equation}
where $w_p$ is a weighting factor for each pair (column) which we set to the difference in ranks of the documents represented by $\x_{i}$ and $\x_{j}$. Intuitively, pairs with larger rank difference are more important to preserve than smaller ones in order to maximize $\tau$ and hence are given higher weights. Initially when $\f'$ is empty, $z^{f}_{p}$ is impact of a single feature $f$ on the pairwise concordance of $p \in P$.

\mpara{Utility Computation in Greedy Incremental Search} We utilize a greedy iterative procedure to select features. In the first iteration $\f'$ is empty and $\f' \cup f = f$. We compute the preference matrix as described above for all features $f \in F$ and all pairs $p \in P$. In order to select $k$ features to populate $\f'$,
we estimate the utility of adding a feature to $\f'$ by aggregating the likelihood values for every pair in $P$, i.e. 
$u_i(f,P) = \sum_{p \in P} z^{f}_{p},$
where $u_i(f,P)$ denotes the utility of adding feature $f$ in the $i$th iteration. We select the feature $f$ that has the maximum utility in this iteration to be added to $\f'$. Once the feature $f$ is selected ($\f' = \f' \cup f$), we remove the row corresponding to $f$ from the preference matrix. We then recompute the preference matrix with the new $\f'$. In step $i+i$, we once again estimate the utility of all features if added to $ \f'$. As in traditional greedy approaches the stopping criteria is to halt when adding a new feature does not increase utility. Simply put, if $$\argmax_{f \in \f \setminus \f'} u_i(f,P) > \argmax_{f^* \in \f \setminus (\f' \cup f)} u_{i+1}(f^*,P)$$ then do not add $f^*$. This means that an explanation can also be smaller than $k$ if a new feature does not improve utility. We denote this approach in our experiments as \greedy. The pseudocode is given in Algorithm~\ref{alg:incSearch}. We shortened $\mathcal{R}(\q,\f)$ to $\mathcal{R}(\f)$ since $\q$ is implicit in our setting.



\begin{algorithm}[h]
\begin{algorithmic}[1]
\caption{\greedy}
\label{alg:incSearch}
\Require{Trained LTR model $\mathcal{R}$, input feature set $\f$ corresponding to query-document pairs, size of explanation $k$}
\Ensure{Valid Explanation $\f'$ of size $k$}
    \Function{GreedySearch}{$\mathcal{R},\f$}
    \State{Initialize $\f'=\emptyset$}
     \State{Initialize $P$ with sampled concordant(w.r.t. original ranking) pairs}
\For{$(i=0,1,\ldots k-1)  $} 
\State{Choose $f= \argmax {u_i(f,P)} $ }
\State{Let $f^*$ is the feature selected in the $i-1$th iteration}
 \If{$u_i(f,P) > u_{i-1}(f^*,P)$}
               \State{$\f'\leftarrow \f' \cup f$ }
 \EndIf
\EndFor
\EndFunction
\end{algorithmic}
\end{algorithm}

\mpara{Algorithms \greedycov and \greedycovep.} We note that in each iteration, the \greedy  algorithm does not always guarantee an increase in the number of concordant pairs when adding $f$ to the explanation set. Especially when adding a correlated feature, utility might increase when the score difference increases for a pair which was already concordant in the previous step for instance. However, we know that increasing the number of concordant pairs directly leads to an increase in $\tau$. 
To account for such cases, we devise an improved heuristic that instead of maximizing utility over all concordant pairs $P$ at step $i$ only maximizes utility for all pairs $\P \subseteq P$. $\P$ is the set of all pairs that were not already considered concordant by the explanation set at step $i-1$, i.e. we want to improve the concordance propensity of pairs previously considered to have low $z^{f}_{p}$. $P\setminus \P$ is then the set of pairs already ``covered" by $\f'$ until step $i$. 

Just like $\f'$, $\P$ is also updated incrementally. After selecting the feature $f$ with maximum utility at step $i$, all pairs $p$ for which $z^{f}_{p} < \epsilon$ are added to $\P$. $\epsilon$ is a threshold that allows us to control which pairs are considered still unaccounted for. When $\epsilon = 0$ we strictly consider only pairs that were still discordant in the previous iteration. Raising the value of the coverage factor $\epsilon$ lets us also retain pairs for which $z^{f}_{p}>0$ in $i+1$. The intuition here is that a small $z^{f}_{p}$ is a weak indicator of concordance. 

Note that our approach is predicated on the assumption that adding a feature to $\f$ does not change the concordance of a previously covered pair. While this assumption can seem limiting, our experiments show promising results where the improved strategy maximizes validity by focusing on different parts of the ranked list incrementally. Intuitively, the stopping criteria now becomes coverage oriented, i.e. when $\P = P$.

We call this greedy heuristic variation \greedycov  when $\epsilon=0$ and \greedycovep  when $\epsilon>0$ in our experiments. The pseudocode of our algorithm \greedycovep is provided in Algorithm~\ref{alg:greedyep}. 
 %
 \begin{algorithm}[h]
\begin{algorithmic}[1]
\caption{\greedycovep}
\label{alg:greedyep}
\Require{Trained LTR model $\mathcal{R}$, input feature set $\f$ corresponding to query-document pairs of $\q$, size of explanation $k$, coverage factor $\epsilon$}
\Ensure{Valid Explanation $\f'$ of size $k$}
    \Function{\greedycovep}{$\mathcal{R},\f$}
    \State{Initialize $\f'=\emptyset$}
    \State{Initialize $P$ with sampled concordant(w.r.t. original ranking) pairs}
\For{$(i=0,1,\ldots k-1)  $} 
\If{$P\neq \emptyset$}
    \State{Choose $f= \argmax {u_i(f,P)} $ }
    \State{Set $P= P\setminus \{p: z_p^f>\epsilon\}$ }
    \State{$\f'\leftarrow \f' \cup f$ }
\EndIf
\EndFor
\EndFunction
\end{algorithmic}
\end{algorithm}

\subsection{Comparison with SHAP}
\label{sec:compareSHAP}
Kernel \shap{} is the model agnostic version of SHAP~\cite{lundberg2017unified} that we compare our approach against in the experiments. Kernel \shap{} approximates the original model, $f(x)$ on input feature vector $x$ (of size $M$) by a linear explanation model $g(x')$ on binarized input $x'$ such that
$g(x')=\phi_0 + \sum_{i=0}^M \phi_i x'_i,$
where $x'_i\in\{0,1\}$ depending on whether $i$th feature is used or not. $\phi_i$ corresponds to the importance/attribution of feature $i$ for model prediction. This version of SHAP is equivalent to LIME~\cite{ribeiro2016should} with the new kernel mentioned in~\cite{lundberg2017unified} to better estimate the shapley feature attributions.  

Recall that the \ltr{} model takes as input a query-document instance and outputs a relevance score for the document. That means we can use \shap{} for a given query-document input to obtain the corresponding feature importance coefficients for the predictive score of a particular document. One can in principle add the corresponding feature coefficients by running \shap{} for all documents. An explanation set is then computed using the top $k$ features with highest importance coefficients. In our experiments, we observed that it is best to use the importance values for the top ranked document. 

\mpara{Limitations of SHAP.} As opposed to our approach, \shap{} does not directly output an explanation set. Instead it must be inferred from the feature attributions towards the predicted score of each document. \textbf{First}, there is no principled way in which these attributions should be aggregated for explaining scores of a list of documents at the same time. 
\textbf{Second}, \shap{} has a strong independence assumption between features while computing attributions and disregards the feature dependence structure. 
Our solution on the other hand actively considers dependence between features in the selection of feature sets.
So while \shap{} might be accurate in determining feature-wise importance, it fails to accurately determine the importance of sets of features.

\section{Experimental Setup}
\label{sec:setup}

\mpara{Datasets.} We chose two benchmark learning-to-rank datasets from LETOR4.0~\cite{QinL13} to evaluate our approaches. \mq{} consists of 800 queries with labeled query-document feature vectors. 
There are 46 features ranging from text based features like BM25 and network based features like PageRank. 
\mslr{} is a bigger dataset consisting of 10k queries and 136 features per query-document vector. For all experiments we use fold1 of the train-test split out of the 5 provided.


\mpara{Ranking models.} We selected one model from each of the three main learning-to-rank categories to explain -- \pow{}, \paw{} and \liw{} models. We chose the best performing ranking models in terms of their ranking performance (@NDCG scores) for each of the datasets. However, the ranking performance of our trained models has no bearing on our explanation method.

\begin{wraptable}{r}{40mm}

\begin{tabular}{rcc}
\toprule
          & \mq{} & \mslr{} \\
\midrule
\pow{} &0.471    & 0.229     \\
\paw{} &0.468    & 0.336     \\
\liw{} &0.655    & 0.468    
\caption{
\textbf{Ranking performance of trained models (NDCG@10).}
}
\end{tabular}

\label{tab:ndcg_models}
\end{wraptable}

We chose 
\textsc{Linear Regression} as the \pow{} and  \textsc{LambdaMart}~\cite{wu2010adapting} as the \liw{} for both datasets.
Note that \textsc{LambdaMart} is a type of GBDT (Gradient Boosted Decision Tree) that is optimized for NDCG and is a popular choice in most commercial search engines. 
For \mq{}, we chose \textsc{RankNet} ~\cite{burges2005learning} as \paw{}. However for \mslr{} we used \textsc{RankBoost}~\cite{freund2003efficient} instead of \textsc{RankNet} since the ranking performance in NDCG was significantly better ($0.33$ vs $0.20$).
All models were trained using the RankLib~\footnote{https://sourceforge.net/p/lemur/wiki/RankLib/} implementation of the aforementioned algorithms with default parameter settings.




\mpara{Baseline and competitors.} We use a random selection of $k$ features as the baseline for all our experiments called \textsc{Random}. We use \shap{} (see Section~\ref{sec:compareSHAP}) as our baseline in two different variations. The approach named \sone{} uses SHAP's output explanation corresponding to the score of the top ranked document. We also devised an approach called \sfive{} that aggregates the feature attribution from \shap{} for the top 5 ranked documents. The aggregation function we chose is a simple sum of the feature attribution values. To obtain an explanation of size $k$, we select the top-k features as per their aggregated attribution values. Kernel \shap{} has two hyperparameters \texttt{n-samples} and  \texttt{background-size} that we set to 200 and 500 respectively after tuning.


\mpara{Our approach variants.} In the previous section we presented 3 variations of our approach -- \greedy, \greedycov and \greedycovep. For \greedycovep we set the threshold to be the mean of the concordant pair propensity scores for a given feature (i.e. mean of the values in the corresponding row of the preference matrix). Formally, for a selected $f$

$$
\epsilon_{f} = \frac{\sum_{z \in Z} z}{|Z|}
$$

where $Z = \{ z : z^{f}_{p} > 0$  for all $p \in \P\}$. For all greedy approaches we set the number of pairs sampled from $\pi$ to 50 for \mq{} and 100 for \mslr{} since \mq{} has smaller ranked lists.

\mpara{Seed selection.} Greedy approaches tend to be sensitive to the first item selected. In order to account for suboptimal first feature selection, we choose the top 3 candidate/seed features in the first iteration and switch to choosing the best from the second iteration there on. Intuitively this can be seen as running our greedy procedure 3 times with a different first feature added to the solution set in each run. This simple strategy greatly improved the selections as compared to starting from a single choice.


\mpara{Metrics.} We measure the goodness of an approach in terms of validity and completeness as described in Section~\ref{sec:problem}. For \mq{} we take all $156$ queries in the test set whereas for \mslr{} we sample $500$ queries from the test set. Validity and completeness is computed for each query's explanation and then averaged. 




\section{Results}
\label{sec:experiments}
With our experiments we answer the following research questions:
\begin{itemize}
    \item \textbf{RQ I:} How effective is our approach in choosing valid explanations compared to the baselines? Section~\ref{sec:validity})
      
    
    \item \textbf{RQ II:} To what degree are the valid  explanations also \textit{complete }?(Section~\ref{sec:completeness})
    
       \item \textbf{RQ III:} How does the obtained valid explanations compare with the optimal explanation? (Section~\ref{sec:discussion})

\end{itemize}

\begin{table}[ht]

\begin{tabular}{lccccccccccc}

\toprule

                             & \multicolumn{3}{c}{Validity $\uparrow$} &  & & \multicolumn{3}{c}{Completeness $\uparrow$} \\
                             & \pow{}   & \paw{}    & \liw{}     &  & & \pow{}     & \paw{}     & \liw{} \\
\midrule
& \multicolumn{8}{c}{\mq{}} \\ \midrule
Random                       & 0.062   & 0.237   & 0.041    & & & -0.155    & -0.718   & -0.305    \\
\textsc{Shap-1}                        & 0.131   & 0.381   & 0.124    & & & 0.039     & -0.517   & -0.068    \\
\textsc{Shap-5}                        & 0.133   & 0.269   & 0.135    & & & 0.041     & -0.391   & \bf{-0.031}    \\
\midrule
\greedy               & 0.202   & 0.287   & 0.259    & & & 0.041     & \bf{-0.369}   & -0.091    \\
\greedycov         & 0.197   & 0.268   & 0.280   & &  & \bf{0.072}        & -0.371   & -0.096    \\
\greedycovep & \bf{0.299}      & \bf{0.616}   & \bf{0.361} &  &  & -0.013    & -0.623   & -0.162    \\
&  &  &  &  &  &  &  \\
\toprule
& \multicolumn{8}{c}{\mslr{}} \\ \midrule

Random & 0.002 & 0.009 & 0.000 & & & -0.172 & -0.634 & -0.303 \\
\textsc{Shap-1} & 0.006 & 0.050 & 0.023 & & & -0.007 & -0.071 & -0.052 \\
\textsc{Shap-5} & 0.009 & 0.048 & 0.001 & & & 0.000 & -0.067 & -0.051 \\ \midrule
\greedy & 0.007 & 0.060 & 0.045 & & & 0.001 & \bf{-0.057} & \bf{-0.034} \\
\greedycov & \bf{0.071} & 0.094 & \bf{0.081} & & & -0.025 & -0.130 & -0.070 \\
\greedycovep & 0.059 & \bf{0.110} & 0.074 & & & \bf{0.020} & -0.074 & -0.041 \\

\toprule
\end{tabular}

\caption{Results for the \mq{},\mslr{} when k=5. All greedy approaches are statistically significantly better than the baselines for validity measures. Results in bold are the best for each model and significantly better ($p \leq 0.05$) that the closest competitor.}
\label{tab:mq2008}
\end{table}

\subsection{Validity of Explanations}
\label{sec:validity}

In Table~\ref{tab:mq2008} we compare the validity of the baseline approaches to our method for $k=5$. 
In the \mq{} dataset, we observe that both our approach and the SHAP-based feature attribution baselines significantly outperform the \textsc{random} baseline.
Secondly, there is a no significant difference between \sone{} and \sfive{} for \paw{} and \liw{} models.
This suggests that aggregating over multiple documents does not improve validity for feature attribution methods like \shap{}.


\mpara{Greedy vs \shap{}.}
Our simple \greedy approach is already significantly better than both \sone{} and \sfive{}. 
Furthermore, \greedycovep is almost $2 \times$ as good as \sone{} for all models in \mq{}. 
The \mslr{} dataset has nearly four times as many features as \mq{} which translates to lower values overall for the same explanation size although a Kendall's $\tau$ value above 0 already indicates that at least 50\% of item pairs are concordant. 
Nevertheless, we find similar trends. All our approaches significantly outperform \shap{} with \greedycovep often being $2 \time -3 \times$ better than \sone{}. 
This shows that directly optimizing for our validity objective is more desirable than choosing features using feature attribution techniques like \shap{}. Note that all greedy variants that we propose significantly outperform \textsc{Shap}. 

\mpara{Effect of Pair Coverage.}
\textbf{} For \mslr{}, \greedycov consistently outperforms \greedy whereas in \mq{}, it is better than \greedy for \liw{} while remaining competitive in the rest. This shows that augmenting a simple greedy heuristic with a coverage criteria generally leads to good results. We incorporated the notion of coverage in our approach to account for the redundancy of correlated features in the explanation. 
If we choose to cover pairs based on a carefully selected threshold like in \greedycovep we can yield significant gains. By essentially ignoring pairs (removed from the preference matrix iteratively) only where the score difference is positive and large, our method can start to select features that are important for other parts of the ranking $\pi$. Redundant features that contribute to the concordance of more ``confident" covered pairs will now have less utility. 
In \mq{}, for the \paw{} model in particular, using only 5 features we are able to produce a $\pi'$ that is very strongly correlated ($0.616$) with $\pi$. In our experiments we chose a simple heuristic to find the the threshold. This heuristic is not always the right choice (\pow{} for \mslr{}) however. For improved performance, one can also manually tune the threshold parameter. 


\begin{figure}
\begin{floatrow}
\ffigbox{%
  \includegraphics[height=1.9in]{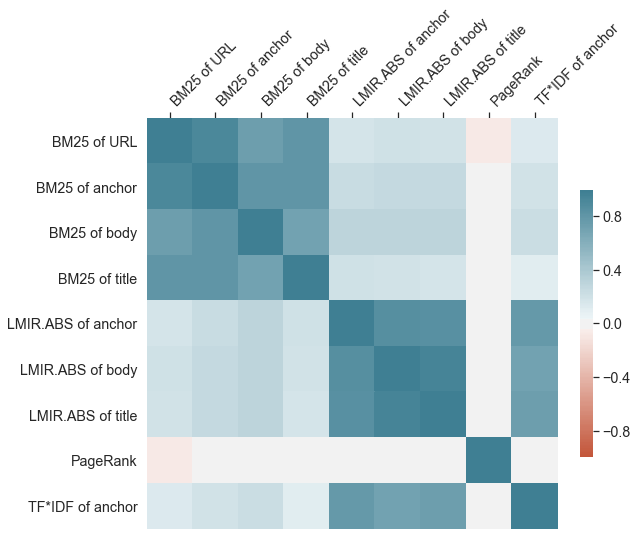}
}{%
  \caption{Feature Correlation of \mq{}}%
  \label{fig:mq_fcor}
}
\capbtabbox{%
 \begin{tabular}{ll|ll}
\toprule
 & \sone & & \greedycovep \\
 \midrule
 & BM25 Body & & BM25 Body \\
 & BM25 Anchor & & TF-IDF Anchor \\
 & LM Anchor &  &BM25 Title \\
 & LM Title & & LM Title \\
 & BM25 URL &  &Page Rank \\
 \midrule
$\mathcal{V}$ & \multicolumn{1}{c}{0.27} & & \multicolumn{1}{c}{1.00} \\
\toprule
\end{tabular}
}{%
  \caption{Explanations for the query \textit{18328} from \mq{} and the \paw{} model}%
  \label{tab:val_ex}
}
\end{floatrow}
\end{figure}

\mpara{Anecdotal example from results.} For query \textit{18328} (from \mq{}) and the \paw{} model, the 5 feature explanation of \sone{} and \greedycovep is shown in Table~\ref{tab:val_ex}.
The validity score of \greedycovep for this query is $1$ while \sone{} is $0.27$. While the features \sone{} selects are good for predicting the relevance of the top ranked document, it is clear that these features alone do not influence the whole ranking as earlier pointed out in Section~\ref{sec:compareSHAP}. Our approach on the other hand is able to explain the entire ranked list by selecting non-redundant holistically (for the list) important features (e.g. PageRank, see Figure~\ref{fig:mq_fcor}).

\subsection{Completeness}
\label{sec:completeness}

\mpara{On the utility of completeness.} We now compare the effectiveness of different explanation methods with respect to completeness that measures the predictive capacity of the features not present in the selected explanation. 
Firstly, we note that while a small number of features may prove to be sufficient for good valid explanations, the same does not hold for completeness. To ensure high completeness all the correlated predictive features need to be selected into the explanation set.
However, longer explanations have negative implications on human decision making as previous works have already established~\cite{lage2019human}.
Neverthless it is indeed important to measure completeness of explanations as a complementary measure to validity in measuring explanation performance.

\mpara{Greedy vs rest.} From Table~\ref{tab:mq2008}, we see that all  approaches are considerably better than a random baseline. 
That is, removing a small random set of features has relatively low impact on the rank order of $\pi'$ (especially for the pairwise model where completeness is $-0.634$). 
Removing 5 important features attributed by \shap{} results in significant gains ($-0.071$ as compared to $-0.634$). Our greedy approaches are better than \shap{} in nearly all cases (except \liw{} for \mq). 
Although, we do not directly optimize for completeness the \greedy approach is able to uncover more complete explanations especially for the \paw{} model. 
For \liw{} models there is however no clear consensus and we observe that the dataset influences whether \shap{} or our approach is a better choice.

\mpara{Why is greedy good?} Out of our approaches, \greedy outputs more complete explanations except for the \pow{} model in \mq{} where it is second best overall. 
\greedy does not take redundancy of feature contributions into account since coverage of pairs in not modelled here. 
Hence it is more likely to select correlated features as compared to \greedycov or \greedycovep reducing its informativeness and validity.

\mpara{Anecdotal example.} \greedy selects the following features as an explanation for query $823$ (from \mslr{}) and the \paw{} ranker -- (i) BM25 score of the body, (ii) LM score of the body, (iii) covered query term ratio of the title, (iv) sum of stream length normalized term frequency for the URL and (v) covered query term number for the title whereas \greedycovep selects (a) LM of the whole doc, (b) sum of stream length normalized term frequency for the URL and (c) covered query term number for the body. The LM and BM25 score of the body are highly correlated but only the LM score is selected by \greedycovep. This results in a completeness score of $-0.1$ for \greedy and $-0.6$ for \greedycovep.


A key takeaway from measuring completeness is that maximizing validity does not always ensure minimizing completeness. 
This points to the presence of small feature subsets that are both valid and complete. 
Finding an explanation that is both valid and complete is a challenging problem due to the dual nature of the objectives (one minimizes rank correlation while the other maximizes it) and we leave this to future work.

\subsection{Comparison to optimal explanations}
\label{sec:optimal}
Now we compare our explanation sets to the one with highest validity score obtained by brute force search over all possible feature subsets of size $k=3$.
We find that for the \paw{} model for \textsc{MQ2008} the validity of the ideal feature set is $0.860$ on average. 
Our best approach produces an explanation $\f'$ (k=5) that achieves an average validity of $0.616$ which shows that even if we make sub-optimal choices in the beginning we are able to exploit pairwise preferences effectively to eventually construct an explanation that is closer to the ideal than all the other approaches. For \pow{} and \liw{} models, the ideal validity is $0.80$ and $0.76$ respectively. Here we see a larger scope for improvement. For \textsc{MSLR}, the ideal validity for all models was approximately $0.3$ which once again shows that we are not too far from the ideal explanation especially for pairwise models ($0.110$ for \greedycovep). 


\begin{figure*}[t!]
    
    \centering
    \begin{subfigure}[t]{0.47\textwidth}
    
        \centering
        \includegraphics[height=1.6in]{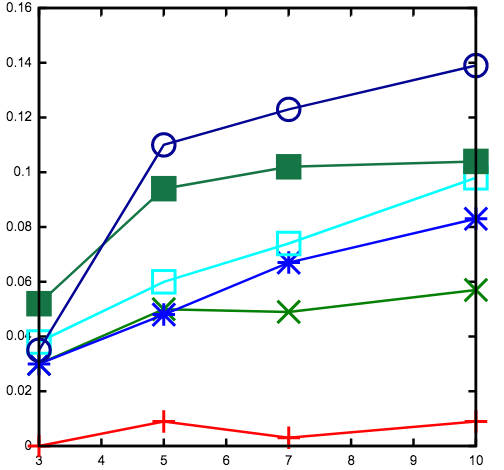}
        \caption{Pairwise Ranker}
        \label{fig:k_pair}
    \end{subfigure}%
    ~ 
    \begin{subfigure}[t]{0.47\textwidth}
    
        \centering
        \includegraphics[height=1.6in]{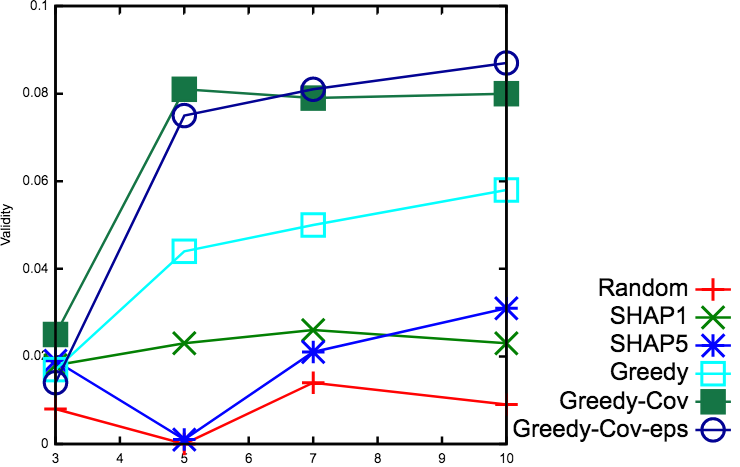}
        \caption{Listwise Ranker}
        \label{fig:k_list}
    \end{subfigure}
    \caption{Effect of $k$. The graphs show the validity of increasing explanation sizes $(3,5,7,10)$ for \textsc{MSLR}.}
    \label{fig:k_graph}
\end{figure*}

\subsection{Effect of k}
\label{sec:discussion}

In section~\ref{sec:problem}, we outlined that our optimization problem was not exactly submodular. Inspite of this, our greedy approaches are still able to find increasingly valid explanations as the size of the explanation is increased. Figure~\ref{fig:k_graph} illustrates this finding for 2 models for \textsc{MSLR}. We found similar trends for the other dataset and models. An interesting finding here is that increasing $k$ benefits \greedycovep more than \greedycov. \greedycov is more likely to be affected by smaller ranked lists (fewer concordant pairs sampled) since pairs will be covered at a faster rate. \greedycovep on the other hand is able to iteratively cover only pairs where it is most confident. By having more pairs in each iteration, the utility of the feature added to the explanation can be better estimated.






\section{Conclusion and Future Work}

In this paper we introduce the novel problem of finding explanations for LTR models in a local posthoc manner. We defined notions of validity and completeness specifically for rankings. We proposed a flexible framework for valid explanations that effectively explores the search space by optimizing for pairwise preference (extracted from the target ranked list) coverage. In our experiments on several model families and datasets we show that our approach is significantly better for LTR models in terms of validity. By comparing against the ideal feature attribution we see that our approach is the first step in this novel problem domain. With the promise of \greedycovep, we envisage a more adaptive and optimization specific threshold leading to improved performance. Finally, we will explore how our approach can be adapted in a semi-blackbox setting where we have access to the learning algorithm but not the model parameters.

\balance
\bibliographystyle{splncs04}
\bibliography{references2}

\end{document}